\documentclass[10pt,conference,final,twocolumn]{IEEEtran}

\usepackage{amsmath,amssymb,graphicx,xcolor}
\usepackage{algorithm,algpseudocode,multirow}
\usepackage{epsfig,epstopdf}
\usepackage{subcaption}
\usepackage[font=small,labelfont=bf]{caption}

\definecolor{myred}{rgb}{0.7, 0, 0}
\definecolor{mygreen}{rgb}{0, 0.7, 0}

\newcommand{\mb}[1]{\mathbf{#1}}

\title{Accelerating deep neural networks for efficient scene understanding in automotive cyber-physical systems}

\author{
\IEEEauthorblockN{Stavros Nousias$^{1}$, Erion Vasilis Pikoulis$^{1,2}$, Christos Mavrokefalidis$^{1,2}$, Aris S. Lalos$^{1}$}
\IEEEauthorblockA{$^1$Industrial Systems Institute, Athena Research Center, Patras Science Park, Greece\\
$^2$Computer Engineering and Informatics Dept., University of Patras, Greece\\
Emails: nousias@isi.gr, \{pikoulis,maurokef\}@ceid.upatras.gr, lalos@isi.gr
}
}

\begin{document}

\maketitle
 
\begin{abstract}
Automotive Cyber-Physical Systems (ACPS) have attracted a significant amount of interest in the past few decades, while one of the most critical operations in these systems is the perception of the environment. Deep learning and, especially, the use of Deep Neural Networks (DNNs) provides impressive results in analyzing and understanding complex and dynamic scenes from visual data. The prediction horizons  for  those  perception  systems  are  very  short and inference must often be performed in real time, stressing the need of transforming the original large pre-trained networks into  new  smaller models, by utilizing  Model Compression  and  Acceleration (MCA) techniques. Our goal in this work is to investigate best  practices for  appropriately  applying novel weight sharing techniques, optimizing the available variables and the training procedures towards the significant acceleration of widely adopted DNNs. Extensive evaluation studies carried out using various state-of-the-art DNN models in object detection and tracking experiments, provide details about  the  type  of  errors that  manifest  after  the  application  of  weight  sharing techniques, resulting in significant acceleration gains with negligible accuracy losses.
\end{abstract}

\section{Introduction}

In recent years, Cyber-Physical Systems (CPSs) play an important role in modern technology \cite{zanero2017cyber},  by interconnecting  computational and physical resources. CPSs are realized by embedded computers and communication networks that govern physical actuators that operate in the physical world, while receiving inputs from sensors, thus creating a smart control loop capable of adaptation, autonomy, and improved efficiency. CPSs have impacted almost all aspects of our daily life connected with, for instance, transportation systems, health-care devices, household appliances, electrical power grids, oil and natural gas distribution and many more. Specifically, in the field of Intelligent Transportation System (ITS) \cite{5959985}, the use of CPSs can lead to comprehensive systems that combine advanced technologies with conventional transportation infrastructures, improving the performance of transportation systems, enhancing travel security, fuel economy and, ultimately, enhancing the travel experience of road users.

One of the most essential operations executed at the ITS CPSs to enable the aforementioned benefits is the perception and understanding of dynamic and complex environments from multi-modal sensor data.  The critical nature of the perception ability \cite{zhang2020empowering} in safety functions for autonomous driving is self-evident: a deviation of, for example, 30 $cm$ in the estimated lateral position of the autonomous vehicle can make all the difference between a “correct” and an “incorrect” (and, potentially, life-threatening) maneuver initiation. One of the major challenges to be addressed, regards the highly dynamic behavior of road users (e.g., pedestrians, cyclists, cars), which can change their motion style in an instance, or start/stop moving abruptly. Consequently, prediction horizons for active perception systems are typically short; even so, small performance improvements can produce tangible benefits. For example, accident analyses \cite{lenard2011typical} show that being able to initiate emergency braking 0.16 $s$ (i.e. five frames at 33 Hz) earlier, at a time to collision of 0.66 $s$, reduces the chance of incurring injury requiring a hospital stay from 50\% to 35\%, given an initial vehicle speed of 50 $km/h$. The aforementioned facts clearly indicate the need for fast and effective scene understanding solutions including, among others, image classification, object detection, object tracking and semantic segmentation. 

Here, we focus on DNN-based object detection and, in particular, on the application of MCA techniques on high-performance, pre-trained detectors. Employing MCA techniques can be critical for the efficient execution of the relevant deep models on embedded devices that are deployed on autonomous vehicles. In the following, first, the positioning of the paper is provided through the description of the relevant bibliography and its contribution. Then, the MCA techniques and the object detectors that are adopted for this study, are briefly described. Finally, before concluding the paper, a thorough experimental evaluation of the MCA impact on the behaviour of the adopted models is presented. 

\section{Relevant bibliography and contribution}

Object detection has been evolved considerably since the appearance of deep convolutional neural networks \cite{zhao2019object}.
Nowadays, there are two main branches of proposed techniques. In the first one, the object detectors, using two stages, 
generate region proposals
which are subsequently classified in the categories that are determined by the application at hand (e.g., vehicles, cyclists and
pedestrians, in the case of autonomous driving). Some important, representative, high performance examples of this first branch are 
Faster R-CNN \cite{ren2016faster}, Region-based Fully Convolutional Network (R-FCN) \cite{dai2016r}, Feature Pyramid Network (FPN) \cite{lin2017feature} and Mask R-CNN \cite{he2017mask}. In the second branch, object detection is cast to a single-stage, regression-like
task with the aim to provide directly both the locations and the categories of the detected objects. Notable examples, here,
are Single Shot MultiBox Detector (SSD) \cite{liu2016ssd}, SqueezeDet \cite{wu2017squeezedet}, YOLOv3 \cite{redmon2018yolov3} and EfficientDet \cite{tan2020efficientdet}.

Although two-stage detectors demonstrate better performance than the single-stage counterparts, the latter
have lower computational and storage requirements which leads, generally, to faster inference time  \cite{yurtsever2020survey}. 
In autonomous driving, Advanced Driver Assistance Systems (ADAS) rely on embedded systems with limited resources. ADAS is
responsible of executing various machine learning tasks, including object detection, meaning that efficient implementations
that take into account those limitations are critical \cite{borrego2020resource}. To this end, single-stage detectors have been particularly studied for autonomous driving by either
proposing specialized, compact deep models (e.g., \cite{kozlov2019development}, SqueezeDet \cite{wu2017squeezedet}, SA-YOLOv3 \cite{tian2020sa}, Mini-YOLOv3 \cite{mao2019mini}) or applying MCA techniques \cite{Deng2020} to existing, pre-trained models (e.g.,\cite{krittayanawach2019robust} \cite{xu2019training}, \cite{nguyen2020towards}, Efficient YOLO \cite{wang2020efficient}, ICME 2020 Competition \cite{tsai20202020}).


The study and development of methodologies and algorithms for the compression and acceleration of high-performing, yet highly resource demanding deep models, has been a very active area of research in recent years (\cite{Deng2020}, \cite{Sze2017}, \cite{Cheng2018}, \cite{Zhang2019}). 
Three main groups of works have have appeared in the literature.
In the first one, the proposed techniques transform the models by removing / pruning parts of the neural network (e.g., parameters, connections, channels, etc. \cite{Li2016}, \cite{Lin2019}). In the second group, the representation of the involved parameters is limited via scalar, vector and product quantization, allowing the so-called weight sharing (\cite{Gupta2015}, \cite{Gong2014}). Finally, in the third group, the involved parameters are transformed via appropriate tensor / matrix decompositions that impose low-rankness, sparsity, etc., \cite{Bhattacharya2016}.

Most of the MCA techniques that have been applied for the problem of object recognition (as the ones
mentioned above), belong either to pruning or scalar quantization, which currently are supported by toolboxes like the TensorFlow Model Optimization Toolkit. Here, moving a step further, we focus on more elaborate and high-performing MCA techniques that belong to weight sharing  \cite{Cheng2018_quantized}, \cite{ICMLA20} and study their impact on the performance of object detection for autonomous driving. The contributions of the paper are as follows:
\begin{itemize}
    \item Two weight sharing techniques are employed for the compression / acceleration of two object detection
    deep models that are based on the well-known ResNet50 and on SqueezeNet DNNs \cite{wu2017squeezedet}.
    \item An analysis is provided on the error types that manifest after the application of weight sharing techniques.
    \item The results obtained on the KITTI dataset using the selected DNN models, reveal acceleration gains of up to $70 \%$ with negligible accuracy loss.
\end{itemize}

\section{Weight sharing via Product Quantization}\label{sec:problem}

Generally speaking, the linear operation carried out by the convolutional layers can be viewed as involving the computation of dot-products between input and kernel vectors lying in an $N$-dimensional space, with $N$ being the number of input/kernel channels. 


Product quantization first partitions the original $N$-dimensional vector space into a number of subspaces and subsequently performs vector quantization in each of them. Specifically, it approximates (represents) the original sub-vectors lying in each subspace using a codebook of limited size. In doing so, product quantization approximates the original dot-products between the input and kernel sub-vectors, by the ones between the input and the representatives/codewords (whose number is much smaller), hence the great potential for acceleration.

Conventionally, vector quantization is achieved by clustering the sub-vectors using the popular $k$-means algorithm (essentially treating them as data points lying in the corresponding subspace), and employing the cluster centroids as the desired codewords. 
%
However, a recently proposed approach that treats the problem from a Dictionary Learning perspective, has shown very promising results \cite{ICMLA20}, achieving up to 100 $\%$ (or, $2\times$) acceleration gain over conventional techniques, on state-of-the-art pre-trained models (VGG, ResNet, SqueezeNet) from the ImageNet competition. 
%

The new approach presented in \cite{ICMLA20} enables the use of a codebook that is several times larger than the ones obtained via $k$-means-based approaches (for the same target acceleration), which leads to considerable improvement regarding the incurred the quantization error. This is achieved by imposing a special structure to the learned codewords, using a Dictionary-Learning framework.

More specifically, let us define the conventional approximation scheme (referred to as VQ hereafter) regarding the kernel sub-vectors of a particular subspace, as follows:
\begin{equation}
\mathbf{W}\approx \mathbf{C}\mathbf{\Gamma},
\label{eq:W_approx_CG}
\end{equation}
where $\mb{W}$, $\mb{C}$, denote the matrices holding the original sub-vectors, and the codewords (cluster centroids), respectively, in their columns, while each column of $\mb{\Gamma}$ is essentially a one-hot encoding of the codewords in $\mb{C}$ (i.e. one element is equal to $1$ and all others are equal to $0$). Thus, according to \eqref{eq:W_approx_CG}, each original sub-vector (column of $\mb{W}$) is approximated by exactly one of the $K_{vq}$
codewords (columns of $\mb{C}$).

On the other hand, the newly proposed  approach (referred to as DL hereafter) is based on a different codebook structure, namely: 
\begin{equation}
\mb{W}\approx\mb{D}\mb{\Lambda}\mb{\Gamma},
\label{eq:W_approx_DLG}
\end{equation}
where $\mb{W}$ and $\mb{\Gamma}$ are 
as in \eqref{eq:W_approx_CG}, while $\mb{D}$ and $\mb{\Lambda}$ denote the dictionary of normalized atoms, and the matrix of sparse coefficients, respectively. Specifically, each column of $\mb{\Lambda}$ contains $\rho$ non-zero elements, with $\rho$ being the sparsity level. Thus, according to the DL-based approximation defined in \eqref{eq:W_approx_DLG}, each of the codewords contained in codebook $\mb{D}\mb{\Lambda}$, is obtained as a linear  combination of $\rho$ atoms from $\mb{D}$. The codebook size (columns of $\mb{\Lambda}$) in this approximation is denoted as $K_{dl}$ while the dictionary size (columns of $\mb{D}$) as $L_{dl}$, with $L_{dl}<K_{dl}$.

The main advantage of the DL approach lies in its ability to employ codebooks that are much larger in size than the ones used by the VQ approach, for the same target acceleration. This owes to the linearity of the involved dot-products, and comes as a direct consequence of the special structure of DL-based codebook, namely its decomposition into a dense dictionary $\mb{D}$ and a sparse matrix $\mb{\Lambda}$. To be more specific, this endows it with the ability to increase the size of $\mb{\Lambda}$ (thus increasing the codebook size) while at the same time limiting the dictionary size (thus restricting the number of dense dot-products). Equivalently, this results in significant acceleration gains (compared to the VQ approach) for the same quantization error, as shown in \cite{ICMLA20}.

\section{Application on widely adopted DNN models}

\begin{figure*}
    \centering
    \includegraphics[width=0.7\linewidth]{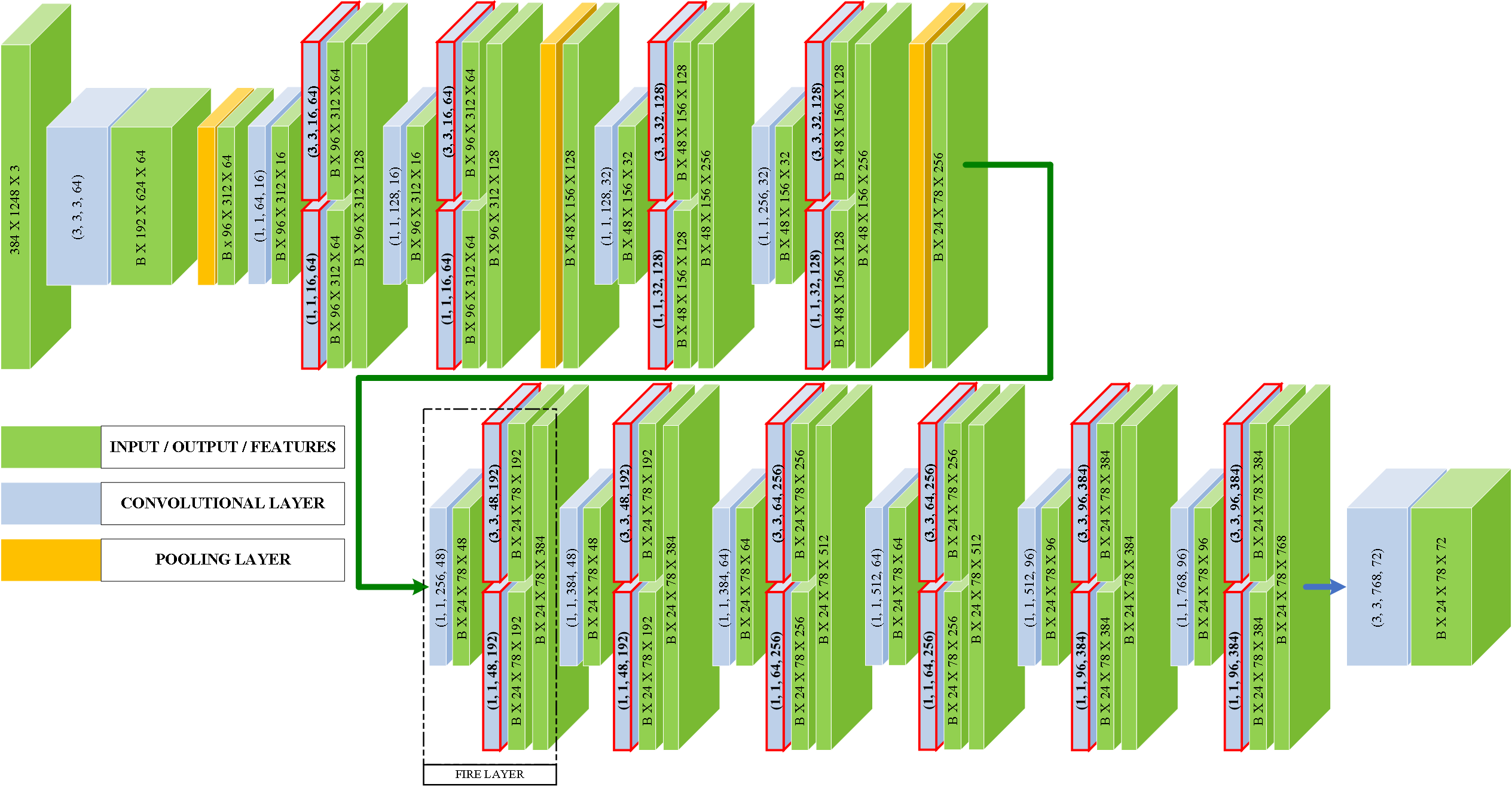}\\
    (a)\\
    \includegraphics[width=0.7\linewidth]{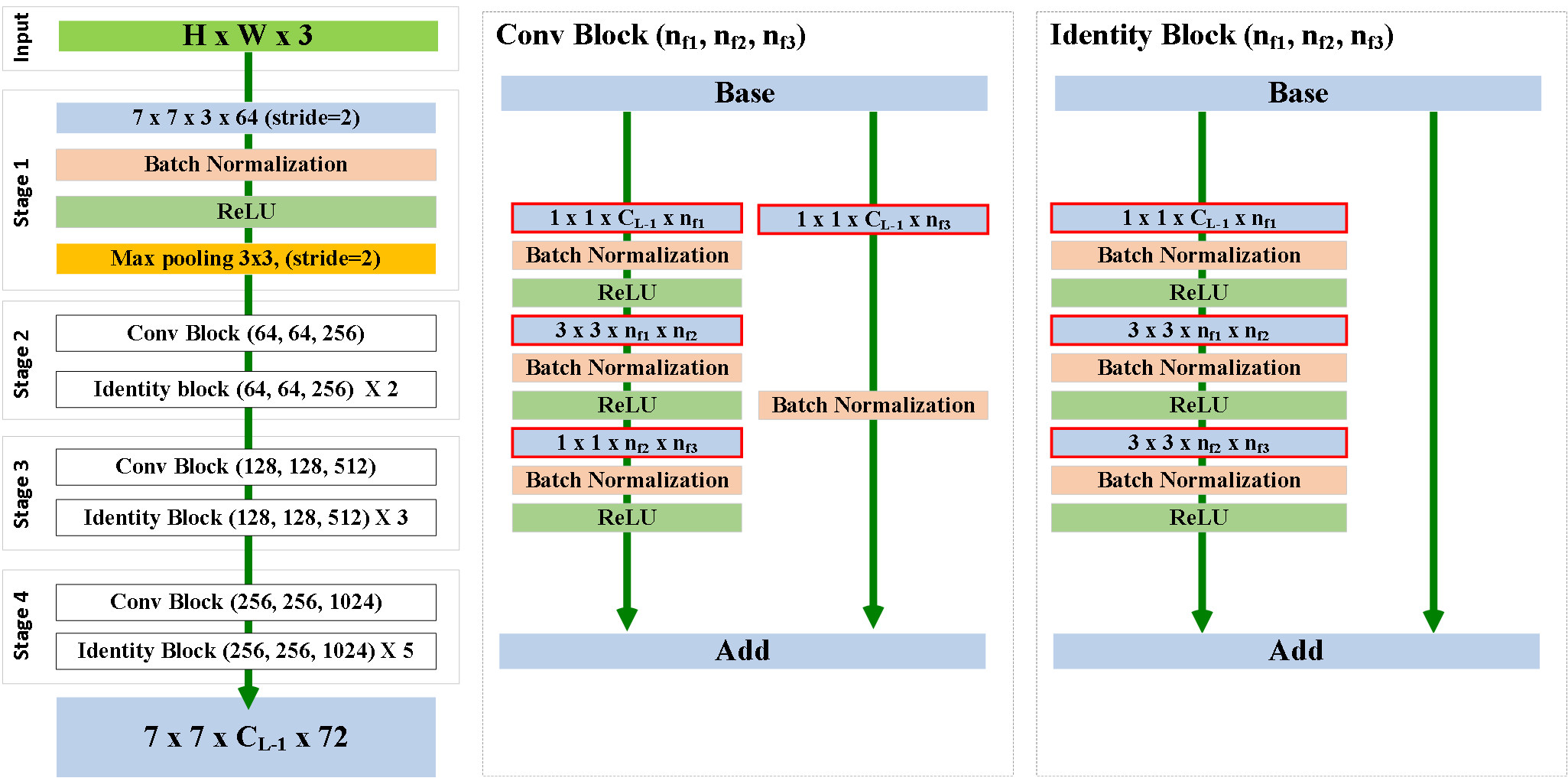}\\
    (b)\\
    \caption{Architectures of the employed detector networks (a) SqueeezeDet, and (b) ResNetDet. The convolutional layers highlighted by the red frames constitute the target layers in our acceleration experiments. $B$ is the batch size, $H$ the height and $W$ the width of a volume kernel. $C_{L-1}$ is the number of channels of the previous layer. }
    \label{fig:networks}
    \vspace{-1em}
\end{figure*}

Two deep detection network architectures, namely SqueezeDet and Resnet50ConvDet, were employed for the evaluation of the presented weight sharing approach. They are fully convolutional detection networks presented by Wu et al. \cite{wu2017squeezedet}, consisting of a feature-extraction part that extracts high dimensional feature maps for the input image, and ConvDet, a convolutional layer to locate objects and predict their class. For the derivation of the final detection, the output is filtered based on a confidence index also extracted by the ConvDet layer. Figure~\ref{fig:networks} presents the overall architecture of the deep networks, the convolutional volume kernel shapes and the feature tensor shapes.

As it can be observed from Fig. \ref{fig:networks}(a), the feature-extraction (convolutional) part of SqueezeDet is based on SqueezeNet \cite{iandola2016squeezenet}, which is a fully convolutional neural network that employs a special architecture that drastically reduces its size while still remaining within the state-of-the-art performance territory. Its building block is the ``fire'' module that consists of a ``squeeze'' $1\times 1$ convolutional layer with the purpose of reducing the number of input channels, followed by  $1\times 1$ and $3\times 3$ ``expand'' convolutional layers that are connected in parallel to the ``squeezed'' output. SqueezeNet consists of $8$ such modules connected in series. 

On the other hand, the backbone of ResNetDet is based on the convolutional layers of ResNet50 \cite{he2016deep}, whose building block consists of three layers, stacked one over the other, as depicted in Fig. \ref{fig:networks}(b). The three layers are $1\times 1$, $3\times 3$, $1\times 1$ convolutions. The $1\times 1$ convolution layers are responsible for reducing and then restoring the dimensions. The $3\times 3$ layer is left as a bottleneck with smaller input/output dimensions. The convolutional part of ResNetDet consists of $13$ such blocks.

\section{Experimental evaluation}

\subsection{Training}
Both networks were trained with the KITTI odometry dataset \cite{geiger2015kitti} consisting of 7477 color traffic scenes images of $1242 \times 375$ pixels. Three classes are taken into account, namely, cyclists, pedestrians and cars which were manually annotated with bounding boxes containing the objects in the scene. A significant observation regarding the dataset is that not all objects of the same class are labeled in each and every image. Such a fact plays a role in the evaluation of the detection outcome as our analysis will reveal. The dataset was split in a $80\%,20\%$ for training and validation, respectively, resulting in $N_{tr}=5980$ training examples and $N_{val}=1497$ validation examples.


For the training of the SqueezeDet architecture, Stochastic Gradient Descent (SGD) was employed with the following values for the hyperparameters (determined via experimentation); batch size $B=8$, learning rate $LR=10^{-4}$, with a weight decay rate $D_{W}=10^{-4}$, a learning rate decay rate of $D_{LR}= 2 *LR / N_e $, number of steps $N_s = 3 \times N_{tr} $ and a dropout rate of $50\%$, over a total of $N_e=300$ epochs. Training and testing took place in an NVIDIA GeForce GTX 1080 graphics card with 8GB VRAM and compute capability $6.1$ in a Intel(R) Core(TM) i7-4790 CPU @ 3.60Hz based system with 32GB of RAM. 

Likewise, for Resnet50ConvDet, we also employed SGD with hyperparameter values as in the case of SqueezeDet. 
Training and evaluation of Resnet50ConvDet took place in an NVIDIA GeForce Geforce RTX 2080 with 16GB VRAM and compute capability $7.5$ in a Intel(R) Core(TM) i7-4790 CPU @ 3.60Hz based system with 16GB of RAM.

In all cases, training took place with a data augmentation scheme where the bounding boxes drift by $k_x*150$ pixels across the x-axis and $k_y*150$ pixels across the y-axis, where $k_x, k_y \sim U(0,1)$. A $50\%$ probability is also assumed to flip the object.

\subsection{Acceleration scheme}

In our experiment, we apply the rival techniques to the two detection models in a ``full-model'' acceleration scenario. It involves accelerating multiple (or all) convolutional layers of the original models and measuring the achieved performance of the accelerated networks. The reported acceleration ratios are defined as the ratio of original vs accelerated computational complexities, measured by required multiply-accumulate (MAC) operations.

The full-model acceleration of the involved networks is achieved by following the progressive strategy proposed in \cite{Cheng2018_quantized}, whereby the individual layers are quantized sequentially in stages, having the original network as a starting point. The quantization operation is followed by a fine-tuning step involving the remaining original layers, after each stage. Fine-tuning and performance evaluation, are based on the training and validation datasets from KITTI, respectively, as previously explained.


\paragraph{Accelerating SqueezeDet}

The feature-extraction part of SqueezeDet, namely SqueezeNet, is responsible for roughly $83\%$ of the total $5.3\times 10^9$ MAC operations required. Since SqueezeNet is specifically designed for efficiency, and in order to maintain a good balance between acceleration and performance, in our experiments we only targeted the ``expand'' layers of the network, as shown in Fig. \ref{fig:networks}(a). Acceleration was performed in $8$ acceleration stages, with each stage involving a particular ``expand'' modules (followed by fine-tuning). Using acceleration ratios of $\alpha=8$, $10$, $12$, and $20$ on the targeted layers, an acceleration of the SqueezeNet part by $72\%$, $74\%$, $75\%$, and $78\%$, respectively, and a total model acceleration by $59\%$, $60\%$, $62\%$, and $65\%$, respectively, were achieved.

\paragraph{Accelerating ResNetDet}

The feature extraction part of ResNetDet is responsible for roughly $81\%$ of the total $3.5\times 10^{10}$ MACs required by the network. Following the network's architecture, in our experiments with ResNetDet, we accelerated its convolutional (feature-extraction) blocks in a one-block-per-stage fashion leading to $13$ total acceleration stages. Using acceleration ratios of $\alpha=8$, $10$, $12$, and $20$ on the targeted layers (see Fig. \ref{fig:networks}(b)), an acceleration of the feature-extraction part by $84\%$, $86\%$, $88\%$, and $92\%$, respectively, and a total model acceleration by $67\%$, $69\%$, $71\%$, and $74\%$, respectively, were achieved.

\subsection{Metrics}

For each detection, the Intersection Over Union (IOU) score is computed as the ratio of area of intersection to the area of union between the predicted and ground-truth bounding boxes. A true positive occurs when IOU$>0.5$ and the predicted class is the same as the ground-truth class. A false positive occurs when IOU$<0.5$ or a different class is detected, meaning that unmatched bounding boxes are taken as false positives for a given class. Precision, recall and mean average precision (mAP) are subsequently calculated according to \cite{powers2020evaluation}.

\subsection{Results}

\begin{figure*}
\begin{minipage}[t]{.245\linewidth}
\centering
\includegraphics[width=\linewidth]{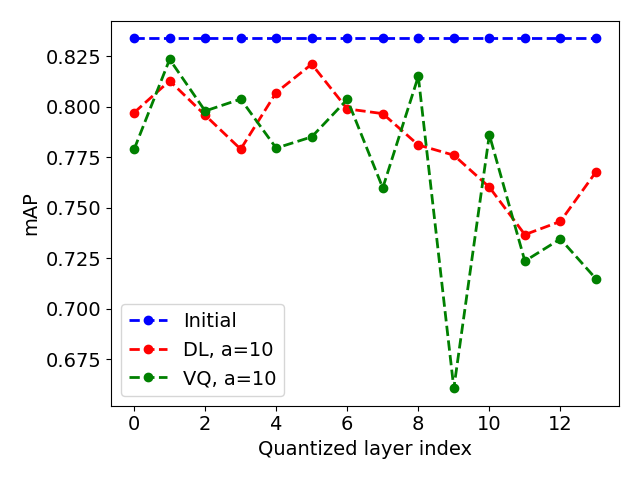}
\centerline{ \scriptsize{(a)}}\medskip
\end{minipage}
\begin{minipage}[t]{.245\linewidth}
\centering
\includegraphics[width=\linewidth]{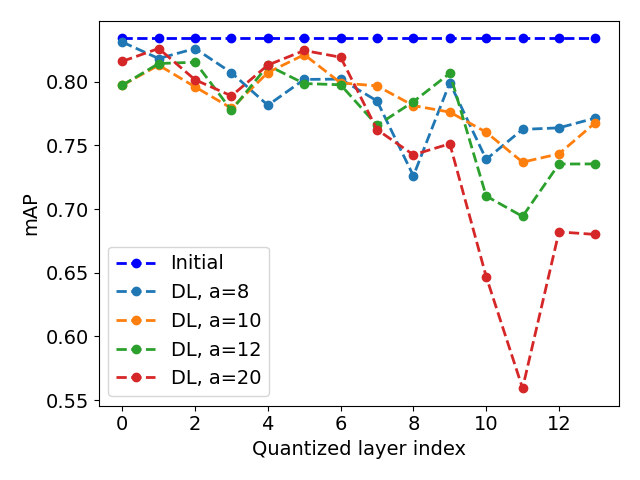}
\centerline{ \scriptsize{(b)}}\medskip
\end{minipage}
\begin{minipage}[t]{.245\linewidth}
\centering
\includegraphics[width=\linewidth]{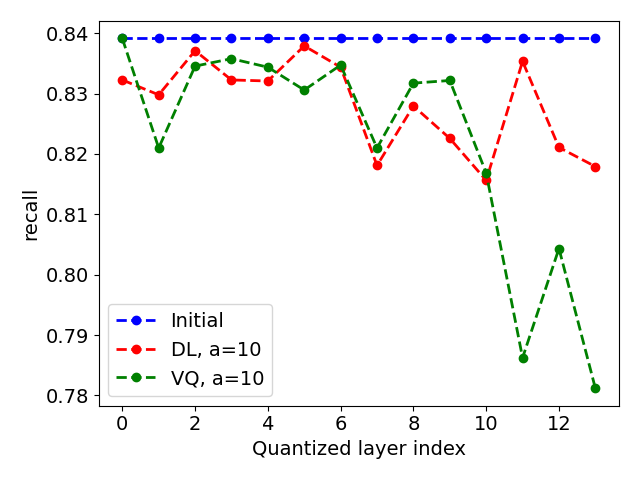}
\centerline{ \scriptsize{(c)}}\medskip
\end{minipage}
\begin{minipage}[t]{.245\linewidth}
\centering
\includegraphics[width=\linewidth]{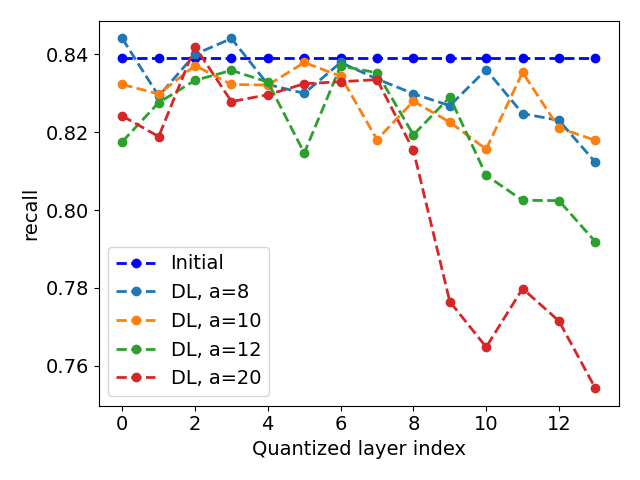}
\centerline{ \scriptsize{(d)}}\medskip
\end{minipage}
\\
\begin{minipage}[t]{.245\linewidth}
\centering
\includegraphics[width=\linewidth]{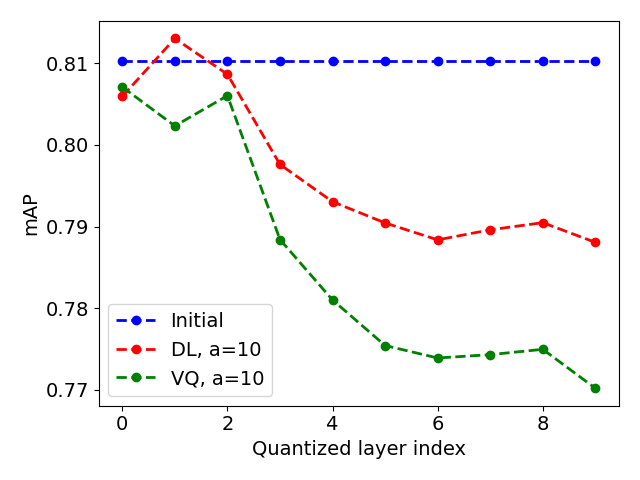}
\centerline{ \scriptsize{(e)}}\medskip
\end{minipage}
\begin{minipage}[t]{.245\linewidth}
\centering
\includegraphics[width=\linewidth]{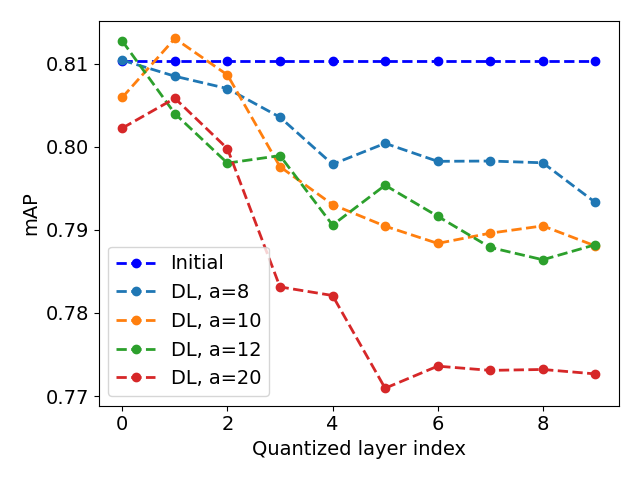}
\centerline{ \scriptsize{(f)}}\medskip
\end{minipage}
\begin{minipage}[t]{.245\linewidth}
\centering
\includegraphics[width=\linewidth]{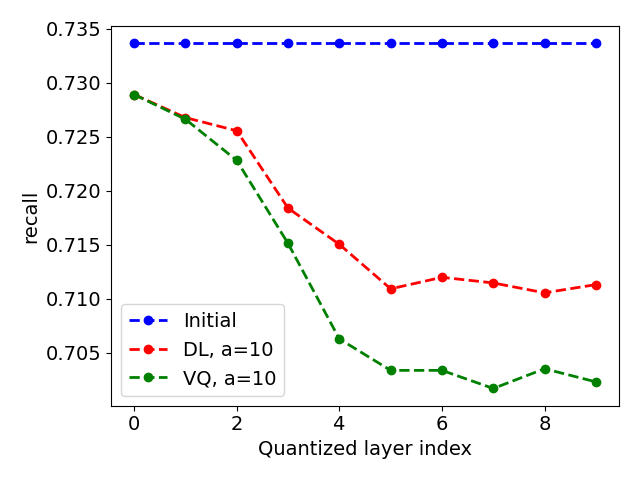}
\centerline{ \scriptsize{(g)}}\medskip
\end{minipage}
\begin{minipage}[t]{.245\linewidth}
\centering
\includegraphics[width=\linewidth]{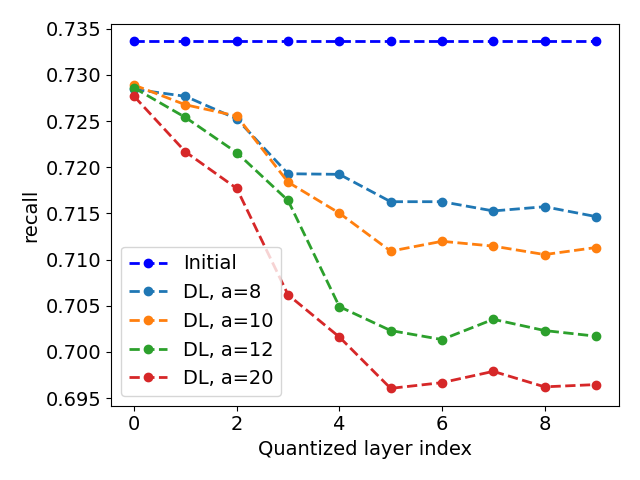}
\centerline{ \scriptsize{(h)}}\medskip
\end{minipage}
\caption{Performance evaluation and comparison of DL vs VQ acceleration techniques on ResNetDet (top row) and SqueezeDet (bottom row).}
\label{fig:progressive_results}
\vspace{-1em}
\end{figure*}

The progressive, stage-wise acceleration results for the employed networks, using both the VQ and the DL acceleration techniques for various acceleration ratios, are shown in Fig. \ref{fig:progressive_results}. The rightmost point in every plot depicts the performance of the ``fully'' accelerated network, i.e., after all targeted convolutional layers have been accelerated. At each point, the performance of the detectors was assessed based on the achieved mean average precision (mAP) and recall. 

As a general comment, the results presented in Fig. \ref{fig:progressive_results} reveal a very promising performance by the employed weight-sharing techniques, and, especially so, for the DL-based one, whose application results in significantly accelerated detectors, with limited loss of their detection capabilities, as expressed by both the mAP and recall values. We stress at this point that the obtained acceleration gains can be further enhanced by better configuring the MCA methodology, so that it is tailored to the specific architecture of the deep model to be transformed. This involves, for instance, careful selection of the layers to be quantized, the amount of compression/acceleration ratio per layer (based on the sensitivity of the layer), etc.
Moreover, comparatively speaking, the DL-based technique managed to generally outperform its rival in our experiments, as highlighted by the plots presented in Fig. \ref{fig:progressive_results}, for an acceleration ratio of $a=10$ (Fig. \ref{fig:progressive_results}(a)$\&$(c), and (e)$\&$(g), for ResNetDet, and SqueezeDet, respectively).

\begin{figure*}
    \centering
    \includegraphics[width=\linewidth]{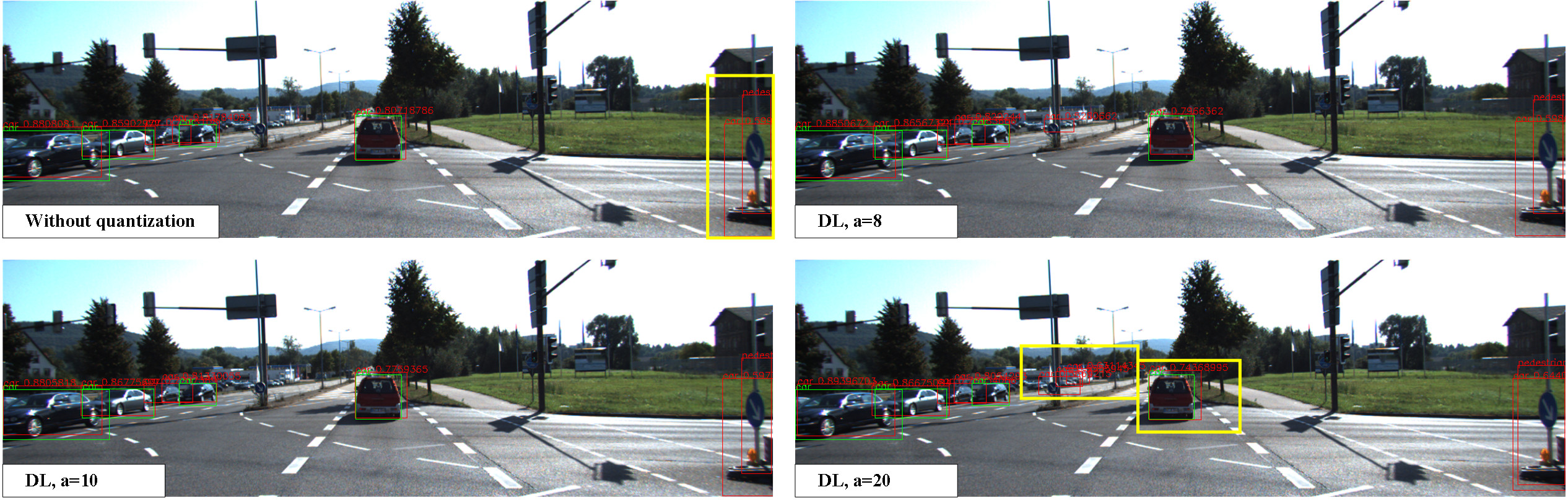}
    \centerline{ \scriptsize{(a) Open street junction}}\medskip
    \\
    \includegraphics[width=\linewidth]{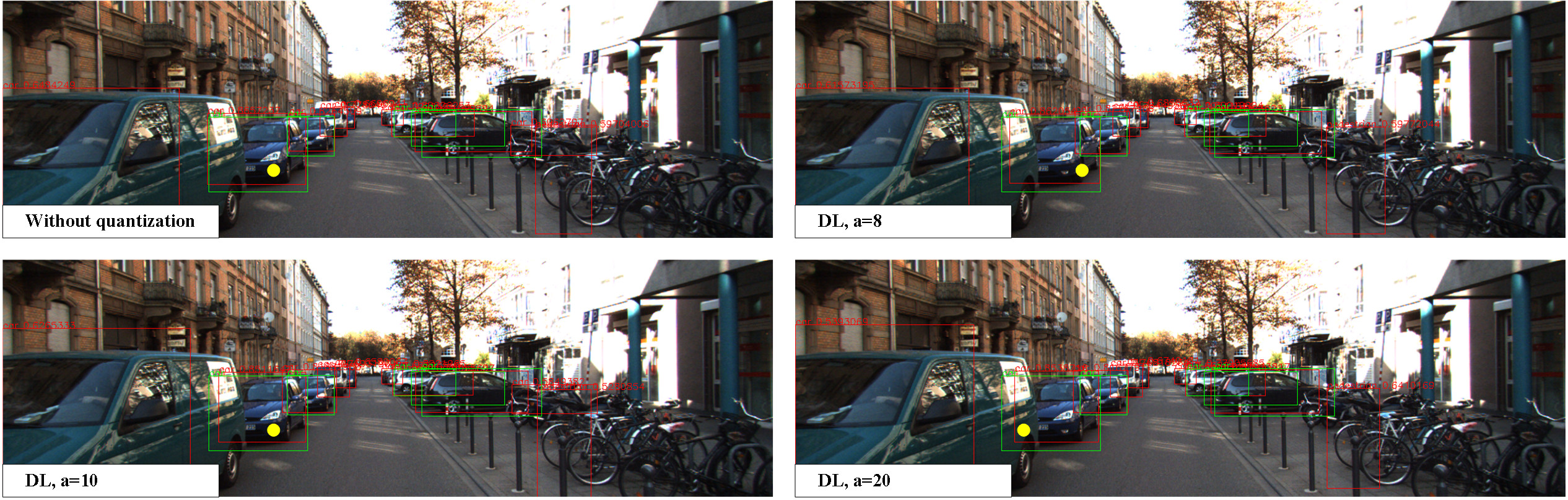}
    \centerline{ \scriptsize{(b) Narrow street}}\medskip
    \caption{Application of accelerated vs original SqueezeDet models, using examples from the KITTI dataset. Green rectangles correspond to ground truth boxes, while red rectangles to predictions. The confidence scores are also shown in red letters. Yellow rectangles in (a) and yellow dot in (b) highlight the most obvious performance degradation of the accelerated networks, as compared to the original one.}
    \label{fig:application_examples}
    \vspace{-1em}
\end{figure*}

Application instances of the accelerated versus the original networks using  examples from the KITI dataset are shown in Fig. \ref{fig:application_examples}, respectively. 

\subsection{Error-type analysis}

\begin{figure}
    \includegraphics[width=\linewidth]{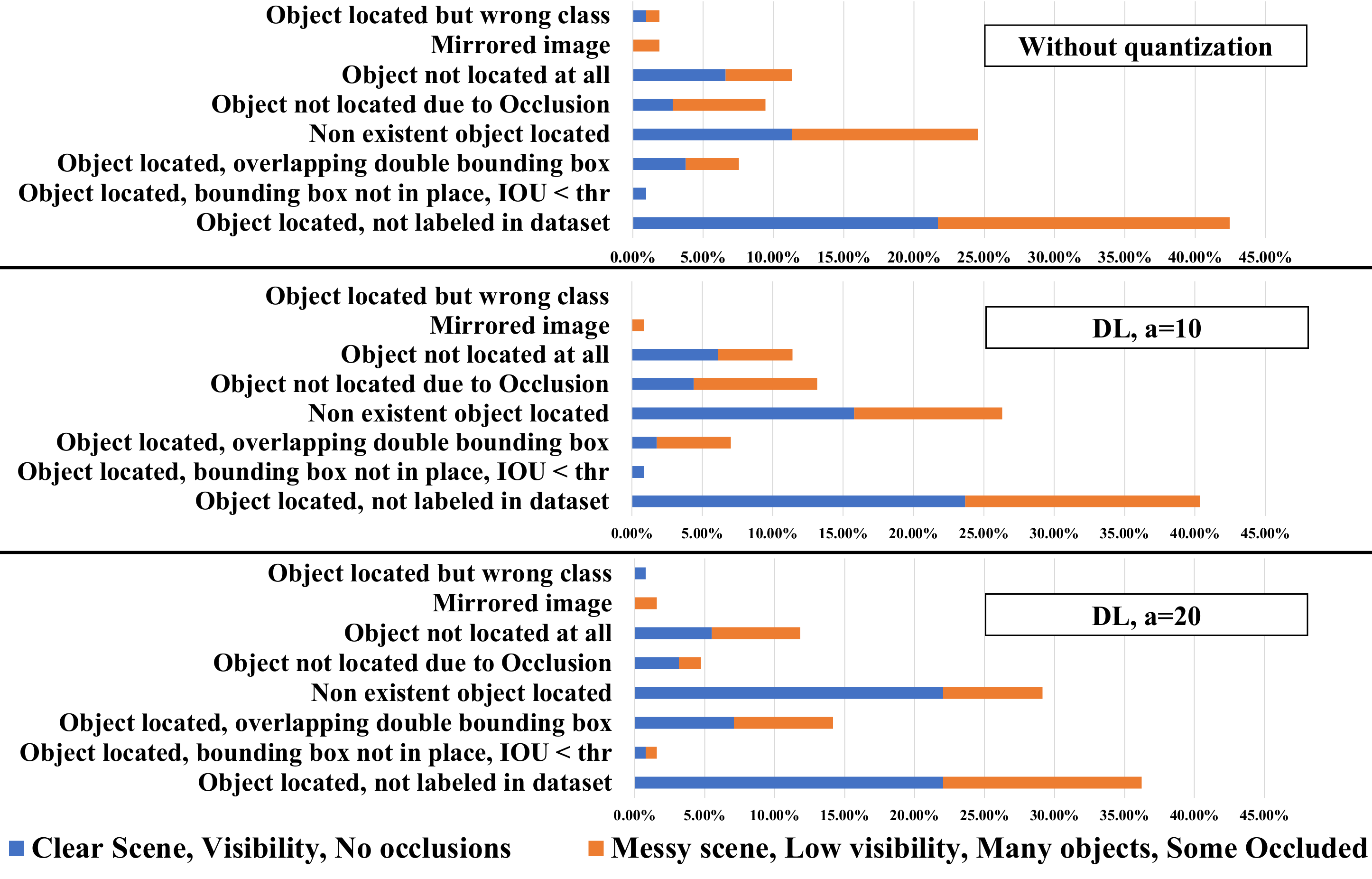}
    \caption{Error type analysis using manually evaluated examples.}
    \label{fig:error_types}
\end{figure}

For a better insight on the obtained results, we performed an in-depth analysis of the error-types of the employed detectors. For this analysis, we examined 64 images containing the groundtruth annotation and the detection outcome and classified the errors into seven categories; a) object located but not labeled in dataset, b) object located buy bounding box not in place ($IOU<0.5$), c) object located but overlapping double bounding box appeared, d) non existent object located, e) object not located due to occlusion, f) object not located at all, and g) mirrored object (i.e., on glass surface), object located but in wrong class. Furthermore, we manually classified the 64 images into clear scenes with sufficient light and no occlusions, and messy scenes with many objects some of them being occluded. The motivation behind this perspective is that the detector correctly detects an object but it is assumed as an error or the detector correctly misses an object (i.e., occlusion) but it is assumed as an error since it was originally annotated in the dataset. As we can observe $50\%$ of the errors in the examined images, are objects that were actually found but either they were not annotated or there was a bounding box issue. The results of this qualitative analysis are summarized by the bar-charts shown in Fig. \ref{fig:error_types}.


\begin{figure}
    \centering
     \begin{subfigure}[t]{\linewidth}
     \includegraphics[width=\linewidth]{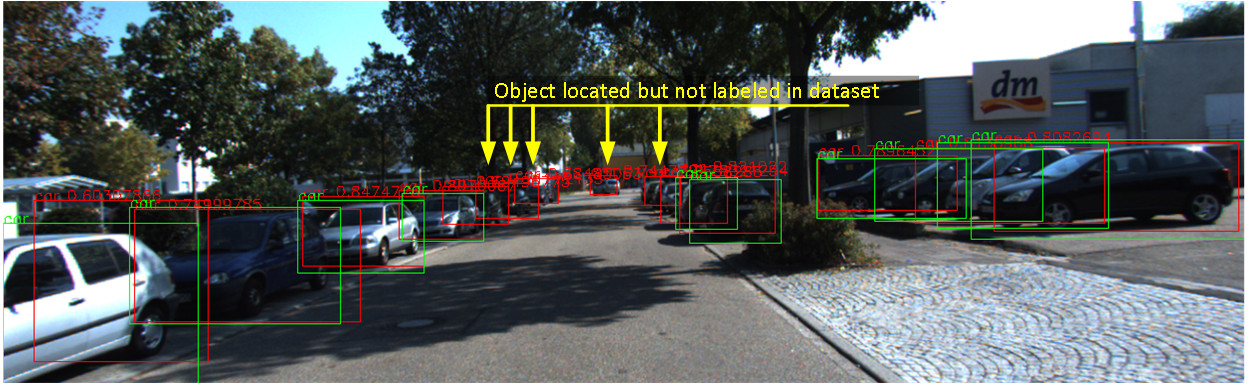}
    \caption{}
    \label{fig:error_case_1}
  \end{subfigure}
  \begin{subfigure}[t]{\linewidth}
     \includegraphics[width=\linewidth]{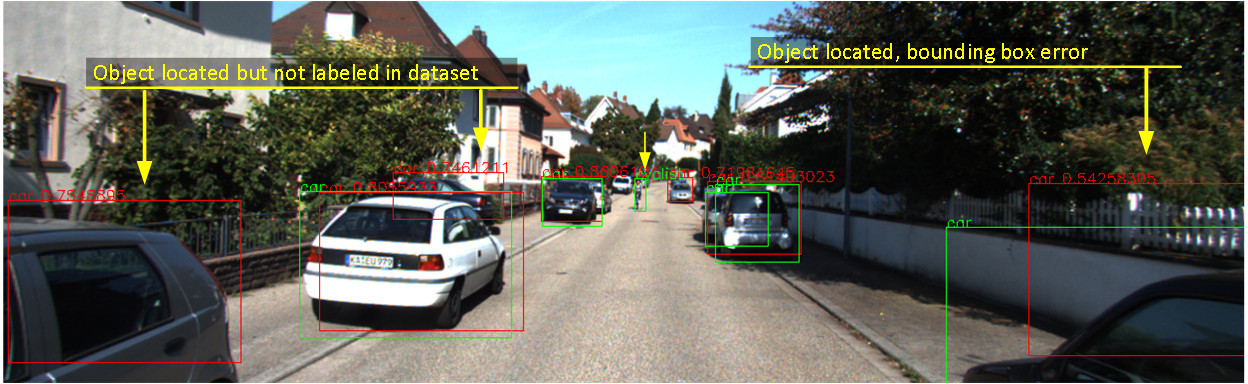}
    \caption{}
    \label{fig:error_case_2}
  \end{subfigure}
  \begin{subfigure}[t]{\linewidth}
     \includegraphics[width=\linewidth]{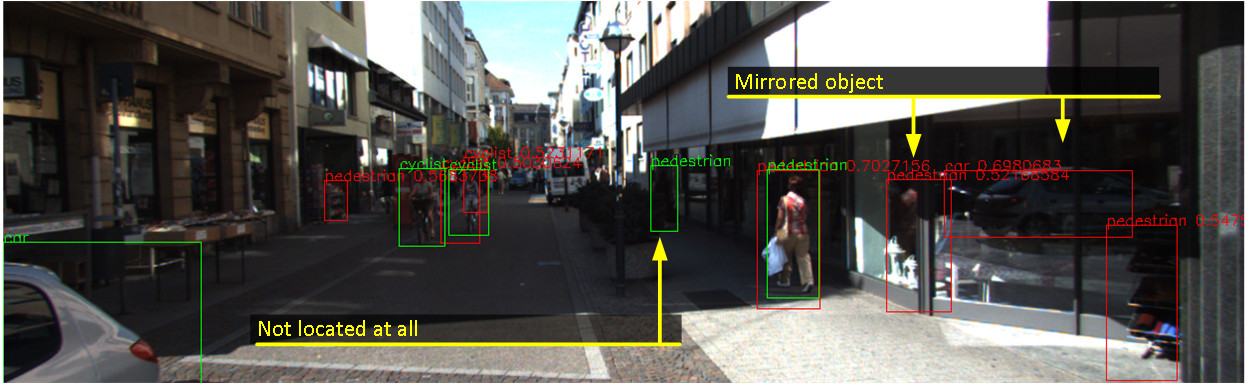}
    \caption{}
    \label{fig:error_case_3}
  \end{subfigure}
  \begin{subfigure}[t]{\linewidth}
     \includegraphics[width=\linewidth]{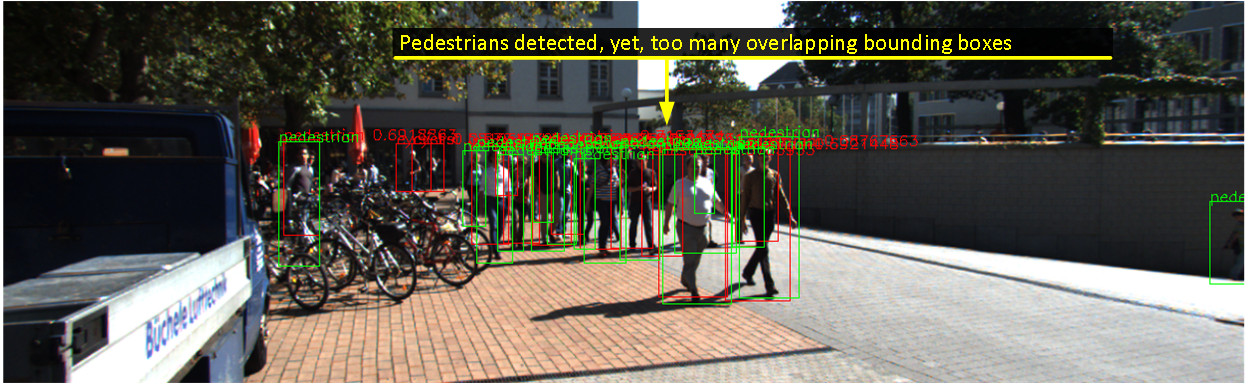}
    \caption{}
    \label{fig:error_case_4}
  \end{subfigure}
    \caption{Examples of error types. Green boxes refer to groundtruth data, while red boxes to detections.}
    \label{fig:error_cases}
    \vspace{-1em}
\end{figure}

\section{Conclusions}

This work investigates the acceleration benefits of weight sharing methods in deep learning based scene analysis for automotive CPSs. Best  practices for optimizing the available variables and the training procedures are described based on extensive evaluations on the KITTI dataset. The presented results provide details about the type  of  errors that manifest, resulting in significant acceleration gains with negligible accuracy losses. By inspecting the error analysis it can be easily seen that most of the errors are attributed to annotation uncertainties. A more thorough investigation that utilizes also synthetic datasets generated from the CARLA autonomous driving simulator is currently under investigation and it is expected to  alleviate the impact of the uncertainties to the training and validation errors, providing additional space for acceleration gains.

\section*{Acknowledgement}

This paper has received funding from H2020 project CPSoSaware (No 873718) and the DEEP-EVIoT - Deep embedded vision using sparse convolutional neural networks project (MIS 5038640) implemented under the Action for the Strategic Development on the Research and Technological Sector, co-financed by national funds through the Operational programme of Western Greece 2014-2020 and European Union funds (European Regional Development Fund).

\label{sec:ref}
\bibliographystyle{IEEEtran}
\bibliography{refs}

\end{document}